%% file: main.tex
\pdfoutput=1

\documentclass[11pt]{article}

\usepackage{acl}

\usepackage{times}
\usepackage{latexsym}
\usepackage{booktabs}
\usepackage{xspace}
\usepackage{tabularx}
\usepackage{enumitem}
\usepackage{graphicx}
\usepackage{cleveref}

\usepackage[T1]{fontenc}

\usepackage[utf8]{inputenc}

\usepackage{microtype}

\title{Automated Metrics for Medical Multi-Document Summarization Disagree with Human Evaluations}

\author{Lucy Lu Wang$^{1,2}$ \quad Yulia Otmakhova$^{3}$ \quad Jay DeYoung$^{4}$ \quad Thinh Hung Truong$^{3}$ \\ \textbf{Bailey E. Kuehl}$^{2}$ \quad \textbf{Erin Bransom}$^{2}$ \quad \textbf{Byron C. Wallace}$^{4}$\vspace{6pt}\\
  $^{1}$University of Washington \quad
  $^{2}$Allen Institute for AI \quad
  $^{3}$University of Melbourne \\
  $^{4}$Northeastern University \\
  \small\texttt{lucylw@uw.edu, \{yotmakhova, hungthinht\}@student.unimelb.edu.au} \\ [-1mm]
  \small\texttt{\{deyoung.j, b.wallace\}@northeastern.edu}
}

\begin{document}
\maketitle

\input{commands.tex}

\begin{abstract}
Evaluating multi-document summarization (MDS) quality is difficult. This is especially true in the case of MDS for biomedical literature reviews, where models must synthesize contradicting evidence reported across different documents. Prior work has shown that rather than performing the task, models may exploit shortcuts that are difficult to detect using standard $n$-gram similarity metrics such as ROUGE. Better automated evaluation metrics are needed, but few resources exist to assess metrics when they are proposed. Therefore, we introduce a dataset of human-assessed summary quality facets and pairwise preferences to encourage and support the development of better automated evaluation methods for literature review MDS. We take advantage of community submissions to the Multi-document Summarization for Literature Review (MSLR) shared task to compile a diverse and representative sample of generated summaries. We analyze how automated summarization evaluation metrics correlate with lexical features of generated summaries, to other automated metrics including several we propose in this work, and to aspects of human-assessed summary quality. We find that not only do automated metrics fail to capture aspects of quality as assessed by humans, in many cases the system rankings produced by these metrics are anti-correlated with rankings according to human annotators.\footnote{Dataset and analysis are available at \githublink.}
\end{abstract}

\section{Introduction}
\label{sec:introduction}
\vspace{-1mm}

\begin{figure}[ht!]
\includegraphics[width=\columnwidth]{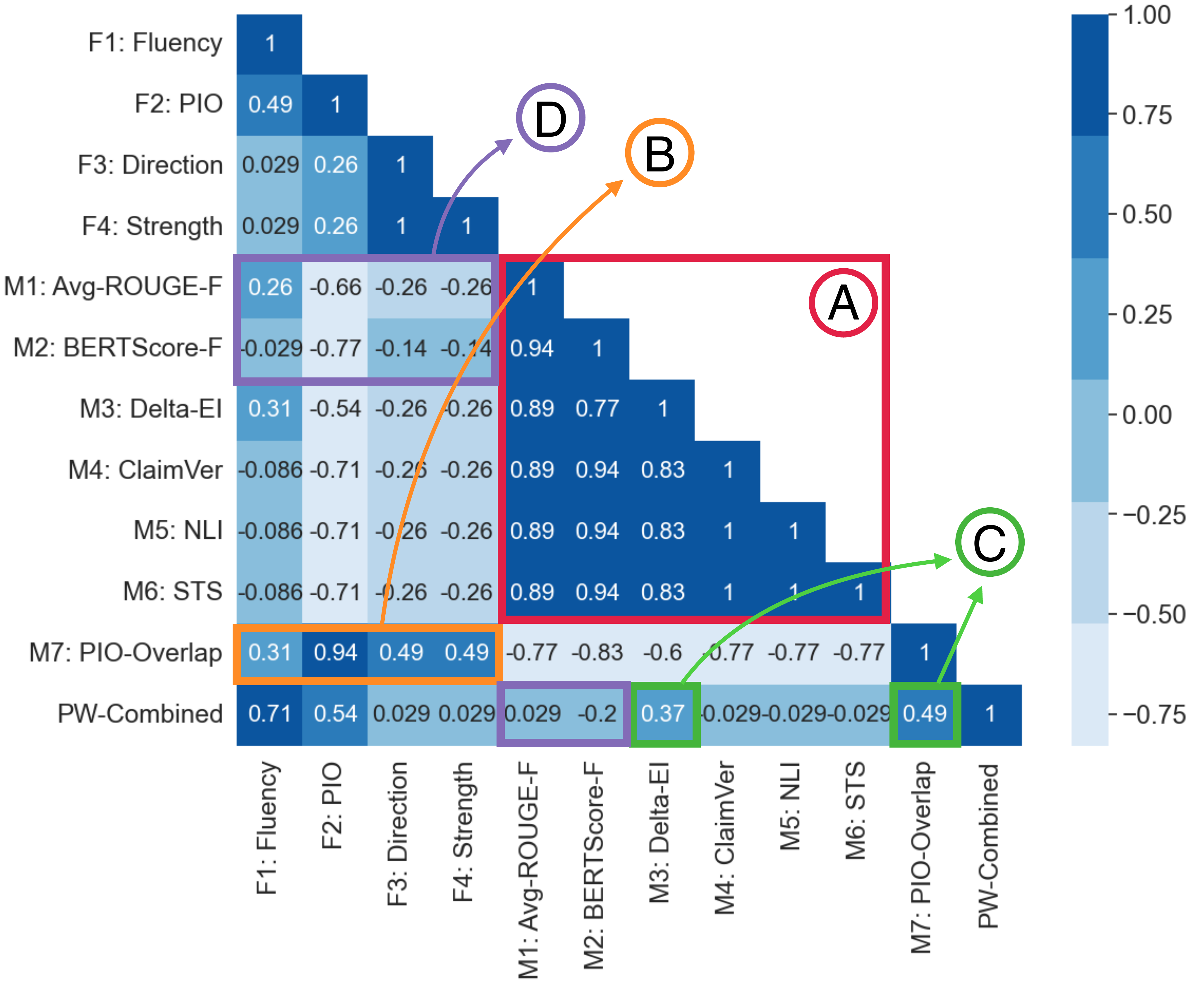}
\caption{Spearman correlations between rankings produced by human-assessed quality facets (F1-F4), automated metrics (M1-M7), and combined pairwise system rankings (PW-combined) on the Cochrane MSLR dataset. Rankings from automated metrics are highly correlated as a group except for PIO-Overlap ({\color{red}{A}}). PIO-Overlap rankings are strongly correlated with rankings from human-assessed facets, especially PIO agreement ({\color{orange}{B}}). Metrics most strongly associated with PW-Combined rankings are Delta-EI and PIO-Overlap ({\color{darkgreen}{C}}). Rankings from commonly reported automated metrics like ROUGE and BERTScore are not correlated or \emph{anti}-correlated with human-assessed system rankings ({\color{violet}{D}}).}
\label{fig:spearman_all}
\end{figure}

Multi-document summarization (MDS) requires models to summarize key points across a set of related documents. Variants of this task have drawn significant attention in recent years, with the introduction of datasets in domains like newswire \cite{fabbri-etal-2019-multi}, Wikipedia \cite{gholipour-ghalandari-etal-2020-large}, science \cite{lu-etal-2020-multi-xscience}, medical literature reviews \cite{DeYoung2021MS2MS, Wallace2020GeneratingN}, and law \cite{Shen2022MultiLexSumRS}; and substantial methodological work to design model architectures tailored to this task \cite{xiao-etal-2022-primera, pasunuru-etal-2021-efficiently, liu-lapata-2019-hierarchical}. 

In this work, we focus on MDS for literature reviews (MSLR), a challenging variant of the task in which one attempts to synthesize all evidence on a given topic.
When manually performed, such reviews usually take teams of experts many months to complete. 
Good review summaries aggregate the results of different studies into a coherent passage, while the evidence presented in the input studies will often be in conflict \cite{Wallace2020GeneratingN, DeYoung2021MS2MS, Wadden2022SciFactOpenTO}, complicating the synthesis task.\footnote{Indeed, reviews conducted by different teams may themselves conflict \cite{ioannidis2016mass}, reflecting the inherent difficulty of the task; however this may owe to differing methods of selecting input studies, a complication we ignore here, though which has been explored in recent work \cite{Giorgi2022ExploringTC}.}

Evaluating conditional text generation models is notoriously difficult, impeding progress in the field. 
Prior work on summarization evaluation has proposed various lexical and modeling-based approaches to assess generation quality, but these metrics predominately use correlation with human-assessed quality facets over relatively small numbers of examples to demonstrate utility \cite{fabbri-etal-2021-summeval, wang-etal-2020-asking, deutsch-roth-2020-sacrerouge, Yuan2021BARTScoreEG}. 
This limitation of current metric evaluation implies that existing automated measures may not generalize well.
Further, evaluation in the multi-document setting adds additional complexity, 
e.g., prior work has shown that MDS models may sometimes exploit shortcuts that do not reflect as detectable changes in automated metrics \cite{Wolhandler2022HowI, Giorgi2022ExploringTC}. 

To address these challenges, we collect 
human annotations to 
evaluate current models and to support automated metrics development for the medical MDS task. 
We construct a dataset of such evaluations using public submissions from the 2022 MSLR shared task on literature review MDS.\footnote{\href{https://github.com/allenai/mslr-shared-task}{https://github.com/allenai/mslr-shared-task}}
Selecting top-performing models, we label the summary quality of a sample of these models' outputs on the Cochrane subtask \cite{Wallace2020GeneratingN}. 
As part of our analysis, we compare system rankings produced by automated metrics and human evaluations. 
Strikingly, our results highlight consistent and significant disagreements between automated metrics and humans, motivating the need for better automated evaluation metrics in this domain.

We contribute the following:

\begin{itemize}[noitemsep, topsep=2pt, leftmargin=10pt]
    \item A dataset of summaries and quality annotations on participant submissions to the MSLR shared task. We include human annotations for 6 models on 8 individual quality facets (\S\ref{sec:facet_annotation}) and pairwise preferences provided by five raters (\S\ref{sec:pw_annotation}). 
    \item An analysis of lexical 
    features among inputs, generated, and target summaries (\S\ref{sec:content}), showing a large amount of undesirable copying behavior.
    \item An analysis of correlations between automated evaluation metrics and human-assessed quality (\S\ref{sec:correlations}), and the differences in system rankings produced by automated metrics versus human evaluation
    (\S\ref{sec:rankings}). We propose several novel evaluation metrics based on desired features of MSLR summaries (\S\ref{sec:correlations}). We find that system rankings derived from commonly reported automated metrics are \emph{not} correlated or even \emph{anti}-correlated with rankings produced by human assessments of quality, though some of the metrics we propose demonstrate promise in capturing certain quality facets.
\end{itemize}

\section{Background}
\label{sec:background}

The MSLR shared task was introduced to bring attention to the challenging task of MDS for literature reviews. 
The shared task comprised two subtasks, based on the Cochrane \cite{Wallace2020GeneratingN} and \mstoo \cite{DeYoung2021MS2MS} datasets. The Cochrane dataset consists of 4.6K reviews from the Cochrane database of systematic reviews. Inputs are abstracts of papers cited by the review and target summaries are the \emph{Authors’ Conclusions} subsections of review abstracts. The \mstoo dataset includes 20K reviews and is semi-automatically constructed from biomedical literature reviews indexed by PubMed. We refer the reader to the original publications for details concerning dataset construction \cite{Wallace2020GeneratingN, DeYoung2021MS2MS}.

Shared task organizers provided training and validation splits for both datasets, and solicited model submissions to two public leaderboards, where models were evaluated on a hidden test split. Models were ranked on the leaderboard using ROUGE (-1, -2, -L; \citealt{lin-2004-rouge}), BERTScore \cite{Zhang2020BERTScoreET}, and Delta-EI \cite{DeYoung2021MS2MS,Wallace2020GeneratingN}, a metric based on evidence inference \cite{lehman2019inferring} classifications. 

\section{Dataset}
\label{sec:dataset}

We construct our dataset from system submissions to the Cochrane subtask leaderboard for the 2022 MSLR shared task (provided to us by task organizers). We only sample from the Cochrane subtask due to the greater number and variety of successful submissions. We include all summaries from the leaderboard, though we only perform human evaluation on summaries generated by 6 models 
(discussion in \S\ref{sec:models}). We define and apply two human evaluation protocols to a sample of summaries from these 6 systems. The first (\S\ref{sec:facet_annotation}) is a facet-based evaluation derived from the analysis conducted in \citet{otmakhova-etal-2022-patient} and the second (\S\ref{sec:pw_annotation}) is a pairwise preference assessment. 

\subsection{MDS systems}
\label{sec:models}

We perform human evaluation on the outputs of 6 MDS systems. Five of these are community submissions to the MSLR-Cochrane leaderboard,\footnote{\href{https://leaderboard.allenai.org/mslr-cochrane/}{https://leaderboard.allenai.org/mslr-cochrane/}} while a sixth is a baseline system (BART-Cochrane) included for reference. These systems 
represent different Transformer model architectures (BART, BART-large, Longformer, BigBird), input selection strategies \citep{Shinde}, and differential representation/attention on input tokens \citep{otmakhova-etal-2022-led, DeYoung2021MS2MS}. We exclude some systems from human evaluation due to 
poor summary quality (disfluent) or being baselines. We briefly describe our 6 systems below.

\paragraph{ITTC-1 / ITTC-2} \citet{otmakhova-etal-2022-led} 
fine-tuned PRIMERA \cite{xiao-etal-2022-primera} for the Cochrane subtask and exploited the use of global attention to highlight 
special entities
and aggregate them across documents. We include two settings from the leaderboard, one that adds global attention to special entity marker tokens (ITTC-1) and one that adds global attention to entity spans (ITTC-2).

\paragraph{BART-large} \citet{Tangsali} fine-tuned BART-large \citep{lewis-etal-2020-bart} for the subtask. 

\paragraph{SciSpace} \citet{Shinde} defined an \emph{extract-then-summarize} approach, combining BERT-based extraction of salient sentences from input documents with a BigBird PEGASUS-based summarization model \cite{Zaheer2020BigBT}.

\paragraph{LED-base-16k} \citet{Giorgi} fine-tuned Longformer Encoder-Decoder \citep{Beltagy2020LongformerTL} for the Cochrane subtask following a similar protocol described in \citet{xiao-etal-2022-primera}.

\paragraph{BART (baseline)} The baseline follows the protocol in \citet{DeYoung2021MS2MS} to fine-tune BART \citep{lewis-etal-2020-bart} for the Cochrane subtask. 

\noindent Model rankings 
originally reported on the MSLR-Cochrane leaderboard are provided in Table~\ref{tab:ranks}.

\begin{center}
\begin{table*}[ht!]
    \footnotesize
    \centering
    \setlength\tabcolsep{4.5pt}
    \begin{tabular}{lccccccc|cccc|c}
        \toprule
        System & ROUGE* & BERTS. & $\Delta$EI & ClaimV. & NLI & STS & PIO-Over. & Flu. & PIO & Dir. & Str. & PW-Comb. \\
        \midrule
        ITTC-1 	 & 5 (4)	 & 5 (2)	 & 4 (6)	 & 4	 & 4	 & 4	 & \cellcolor{darkgreen!25}1	 & 3	 & \cellcolor{darkgreen!25}1	 & 3	 & 3	 & \cellcolor{darkgreen!25}1	\\
        ITTC-2	 & \cellcolor{darkgreen!25}1 (2)	 & 2 (1)	 & \cellcolor{darkgreen!25}1 (2)	 & 2	 & 2	 & 2	 & 5	 & \cellcolor{darkgreen!25}1	 & 4	 & 6	 & 6	 & 2	\\
        BART-large	 & 3 (6)	 & 3 (5)	 & 2 (4)	 & 3	 & 3	 & 3	 & 4	 & 4	 & 5	 & 2	 & 2	 & 3	\\
        LED-base-16k	 & 4 (3)	 & 4 (3)	 & 5 (5)	 & 5	 & 5	 & 5	 & 2	 & 2	 & 2	 & \cellcolor{darkgreen!25}1	 & \cellcolor{darkgreen!25}1	 & 4	\\
        SciSpace	 & 2 (1)	 & \cellcolor{darkgreen!25}1 (6)	 & 3 (3)	 & \cellcolor{darkgreen!25}1	 & \cellcolor{darkgreen!25}1	 & \cellcolor{darkgreen!25}1	 & 6	 & 6	 & 6	 & 4	 & 4	 & 6	\\
        BART (baseline)	 & 6 (5)	 & 6 (4)	 & 6 (1)	 & 6	 & 6	 & 6	 & 3	 & 5	 & 3	 & 5	 & 5	 & 5	\\
        \bottomrule
    \end{tabular}
    \caption{ System rankings based on automated metrics and human evaluation (best in {\color{darkgreen}{green}}). Original system ranks from the MSLR leaderboard as assessed based on ROUGE-L, BERTScore, and Delta-EI are provided in parentheses. The ranks in this table are produced over subsamples of reviews from the Cochrane test split (and macro-averaged for ROUGE and BERTScore), causing ranks to differ from leaderboard rankings. \\ [0.2mm] \footnotesize *Ranking for ROUGE is based on Avg-ROUGE-F, while leaderboard rank is based on ROUGE-L.}
    \label{tab:ranks}
\end{table*}
\end{center}

\subsection{Facet-based Human Evaluation}
\label{sec:facet_annotation}

We adapt a facet-based human evaluation procedure from the analysis in \citet{otmakhova-etal-2022-patient}. In their work, the authors analyzed baseline model outputs from \mstoo \citep{DeYoung2021MS2MS} 
with respect to
fluency, PIO alignment, evidence direction, and modality (or strength of claim). PIO stands for Population (who was studied?~e.g.~women with gestational diabetes), Intervention (what was studied?~e.g.~metformin), and Outcome (what was measured?~e.g.~blood pressure), and is a standard framework for structuring clinical research questions \cite{huang2006evaluation}. 
These are important elements that \emph{must} align between generated and target summaries for the former to be considered accurate. 
Evidence direction describes the 
effect (or lack thereof) that is supported by evidence (e.g., the treatment shows a positive effect, no effect, or a negative effect, comparatively). The strength of the claim indicates how much evidence or how strong the evidence associated with the effect might be. 

We derive 8 questions based on this analysis:

\begin{enumerate}[noitemsep, leftmargin=*, topsep=1mm]
    \item \emph{Fluency}: if the generated summary is fluent
    \item \emph{Population}: whether the population in the generated and target summaries agree
    \item \emph{Intervention}: as above for intervention
    \item \emph{Outcome}: as above for outcome
    \item \emph{Effect-target}: effect direction in the target
    \item \emph{Effect-generated}: effect direction in the generated summary
    \item \emph{Strength-target}: strength of claim in the target
    \item \emph{Strength-generated}: strength of claim in the generated summary
\end{enumerate}
    
\noindent Of the 470 reviews in the Cochrane test set, we sample 100 reviews per system for facet annotations (600 summaries in total). For 50 reviews, we fully annotate all summaries from the 6 systems (the overlapping set); for the other 50 reviews per system, we sample randomly from among the remaining reviews for each system (the random set). All together, at least one system's outputs are annotated for 274 reviews in the test set. We elect for this sampling strategy to balance thoroughness (having sufficient data points to make direct comparisons between systems) and coverage (having annotations across more review topics).

For each sampled instance, we show annotators a pair of (target, generated) summaries from a review and ask them to 
answer 8 questions regarding these (details in App.~\ref{app:facet}). 
A sample of 10 reviews from the overlapping set (60 summary pairs) and 10 from the random set (10 summary pairs) are annotated by two annotators.
We compute inter-annotator agreement from these and report Cohen's Kappa and agreement proportions for all eight facets in Table~\ref{tab:iaa}. Several facets have lower agreement (Population, Outcome, and Strength-target), though most disagreements are between similar classes (e.g.~partial agree vs.~agree); more on this in App.~\ref{app:facet}.

Two annotators with undergraduate biomedical training annotated these samples. We arrived at the final annotation protocol following two rounds of pilot annotations on samples from the \mstoo dataset and discussing among authors to resolve disagreements and achieve consensus.

\begin{table}
\footnotesize
    \begin{tabular}{L{28mm}ccc}
        \toprule
        Question & Classes & $\kappa$ & Agreement \\
        \midrule
        Fluency & 3 & 0.52 & 0.87 \\
        Population & 4 & 0.33 & 0.56 \\
        Intervention & 4 & 0.60 & 0.77 \\
        Outcome & 4 & 0.24 & 0.36 \\
        Effect-target & 4 & 0.85 & 0.90 \\
        Effect-generated & 4 & 0.78 & 0.90 \\
        Strength-target & 4 & 0.30 & 0.54 \\
        Strength-generated & 4 & 0.77 & 0.90 \\
        \bottomrule
    \end{tabular}
    \caption{Inter-annotator agreement between experts on facets (Cohen's $\kappa$ and proportion of agreement).}
    \label{tab:iaa}
\end{table}

\subsection{Pairwise Human Evaluation}
\label{sec:pw_annotation}

We perform pairwise comparisons to elicit human preferences between system-generated summaries and to study how facet-based quality maps to holistic summary quality. 

We sample pairs of system generations from our dataset, half from the overlapping set of reviews annotated for facet evaluations, and half from other reviews. A different subsample of these pairwise comparisons is provided to each of 5 raters, who are asked to complete up to 100 judgments each. For each comparison, the annotator is given the target summary, the system A summary, the system B summary, and asked ``Which of A or B more accurately reflects the content of the target summary?''~where the options are A, B, or Neither. All annotators are knowledgable in BioNLP and one annotator has biomedical training. Four annotators completed 100 pairwise comparisons; a fifth completed 50 comparisons. 

We first determine system rankings per individual annotator. To tally annotations: if A is preferred over B, system A gets 1 point; if B over A, system B gets 1 point; if Neither is preferred, neither system gets a point. Systems are ranked by total points; 
tied systems receive the same ranking.
To determine a combined ranking based on the preferences of all 5 annotators, we adopt the Borda count \cite{Emerson2013TheOB}, a ranked choice vote counting method that maximizes the probability of selecting the Condorcet winner.\footnote{The Condorcet winner is the candidate that would win a head-to-head election against each of the other candidates assuming a plurality vote.} In this method, for each annotator (voter), we award each system the number of points corresponding to the number of systems ranked below it, e.g., for a set of systems ranked 1-6, the rank 1 system receives 5 points, the rank 2 system 4 points, and so on. System rankings resulting from the Borda count are shown in Table~\ref{tab:ranks} under Pairwise-Combined.

We perform bootstrapping over each annotator's pairwise annotations to estimate the error of the overall system rankings. We resample each individual's pairwise preferences with replacement and compute a new combined ranking. Over 10000 bootstrap samples, the average Spearman $\rho$ of the resampled rankings against the initial rankings is 0.716 (s/d = 0.197).

\subsection{Dataset Statistics}

Our final dataset consists of 4658 summaries generated by 10 systems over 470 review instances from MSLR-Cochrane. Of these summaries, 597 from 6 systems are annotated on 8 quality facets. We also include 452 pairwise comparisons from five annotators. In addition to annotations, we compute and include automated metrics for each generated summary to facilitate analysis (more in \S\ref{sec:correlations}).

\section{Analysis of generated summaries}
\label{sec:content}

We perform lexical analysis of input abstracts, system generated summaries, and target summaries in our dataset, summarizing our findings below.

\paragraph{Input copying and synthesis}
To assess similarity between inputs and summaries, we first apply the evidence inference pipeline \cite{lehman2019inferring,deyoung-etal-2020-evidence}\footnote{\href{https://github.com/bwallace/RRnlp}{https://github.com/bwallace/RRnlp}} to identify an evidence statement in each input document and classify it with an effect direction. 
Between each input evidence statement and the target and generated summaries, we compute ROUGE-1 scores. 
We compute the \emph{Synthesis} rate as how often the effect direction agrees between the most similar evidence statement (by ROUGE-1 score) and the generated summary. In Table~\ref{tab:copy_frequency}, we find that system generations match the effect of the closest input at a high rate (0.41-0.46), though no more frequently than we would expect based on the synthesis rate for the target summaries (0.48).
Using ROUGE-1 scores, we also determine how often a generated summary is closer to an input document than the target (\emph{Input Match}), which might indicate whether a system is performing an implicit synthesis by selecting an input and copying it.
We find that systems sometimes copy inputs, but not in any 
consistent way. 

\begin{table}[t!]
\footnotesize

\centering
\begin{tabular}{lcc}

\toprule
System          &  Synthesis       &  Input Match   \\
\midrule
Targets       & 0.48          & -              \\ 
\midrule
ITTC1           & 0.46          & 0.26           \\
ITTC2           & 0.45          & 0.15          \\
BART-Large      & 0.41          & 0.31          \\
LED-Base-16K    & 0.45          & 0.36          \\
SciSpace        & 0.44          & 0.48          \\
BART (baseline) & 0.44          & 0.38           \\

\bottomrule
\end{tabular}
    \caption{Results of summary vs.~input lexical analysis.}
    \label{tab:copy_frequency}
\end{table}

\paragraph{\textit{n}-gram self-repetition}
Previously, \citet{salkar-etal-2022-self} noted that models fine-tuned on the Cochrane corpus tend to generate summaries containing repeating patterns; however, they claim that the amount of such 
\textit{self-repetition}\footnote{\citet{salkar-etal-2022-self} define \textit{self-repetition} as the proportion of generated summaries containing at least one $n$-gram of length $\geq$4 which also occurs in at least one other summary.} is fairly consistent between model-generated and human-written text.
We analyze self-repetition rates for long $n$-grams (5- to 10-grams) and show that their occurrence rates are much higher in generated summaries than in human-written summaries. These long $n$-grams do not just represent stylistic patterns, but can contain important information such as the effect direction, e.g., ``there is insufficient evidence to support the use'' (see App.~\ref{app:self_repetition} for details), so the high rate of self-repetition is very concerning. 

\begin{figure}[t!]
\includegraphics[width=\columnwidth]{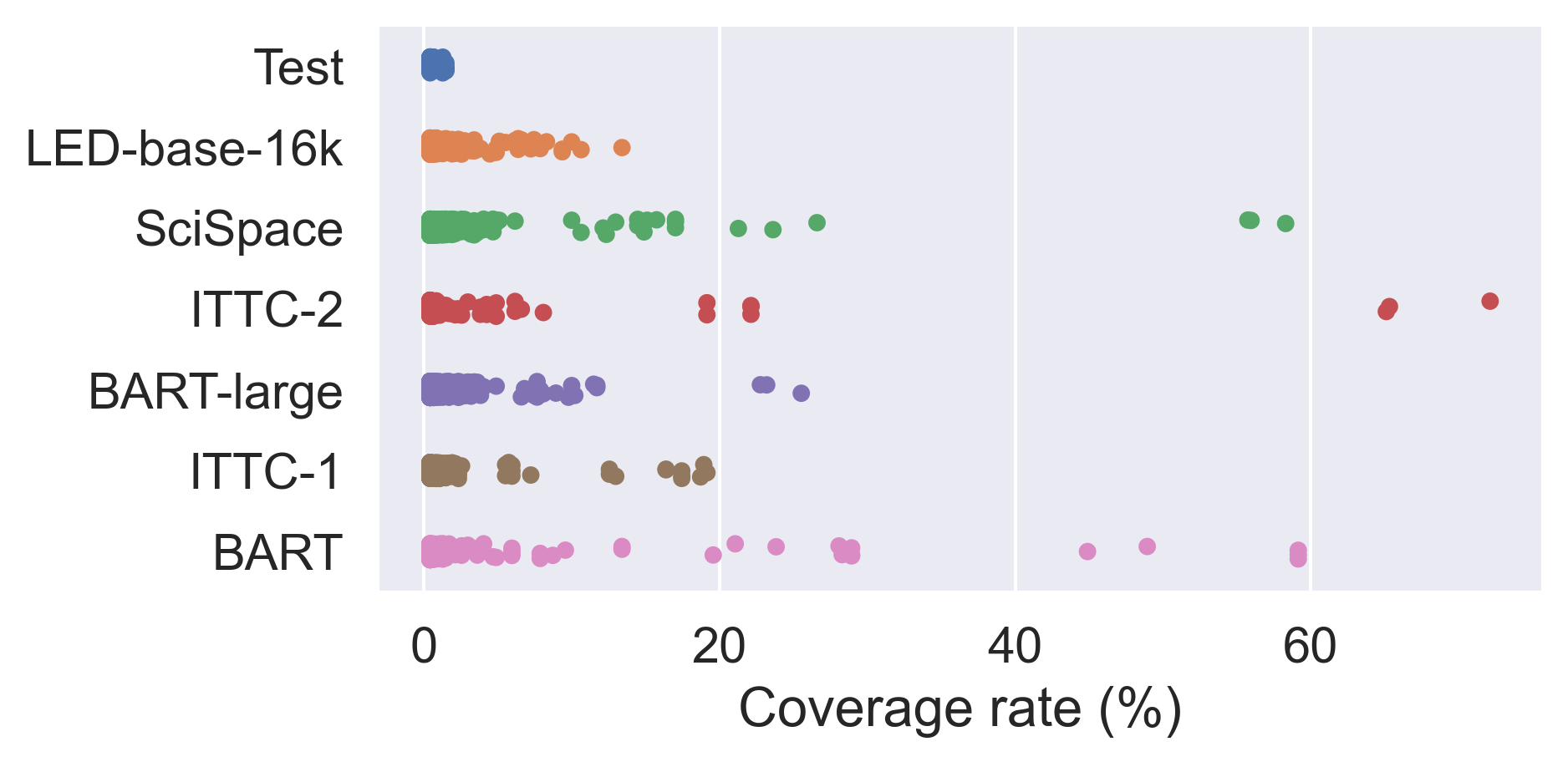}
\caption{Percent of summaries in which each self-repeating 7-gram appears (Test = target summaries).}
\label{fig:ngram_coverage}
\end{figure}

We find a clear distinction between generated and target summaries 
in the self-repetition of longer sequences, such as 7- to 10-grams (Figure~\ref{fig:cochraneselfrep} in App.~\ref{app:self_repetition}). 
Though the amount of self-repeating 10-grams in human-written summaries is negligible, it reaches over 80\% in some of the examined models' outputs. 
The self-repetition rate for specific $n$-grams (the number of documents in which an $n$-gram appears) in generated summaries is also 
much higher than in the targets: some 7-grams occur in up to 70\% of generated summaries (Figure~\ref{fig:ngram_coverage}; trends for other long $n$-grams are in App.~\ref{app:self_repetition}). 

\begin{figure}[t!]
\includegraphics[width=\columnwidth]{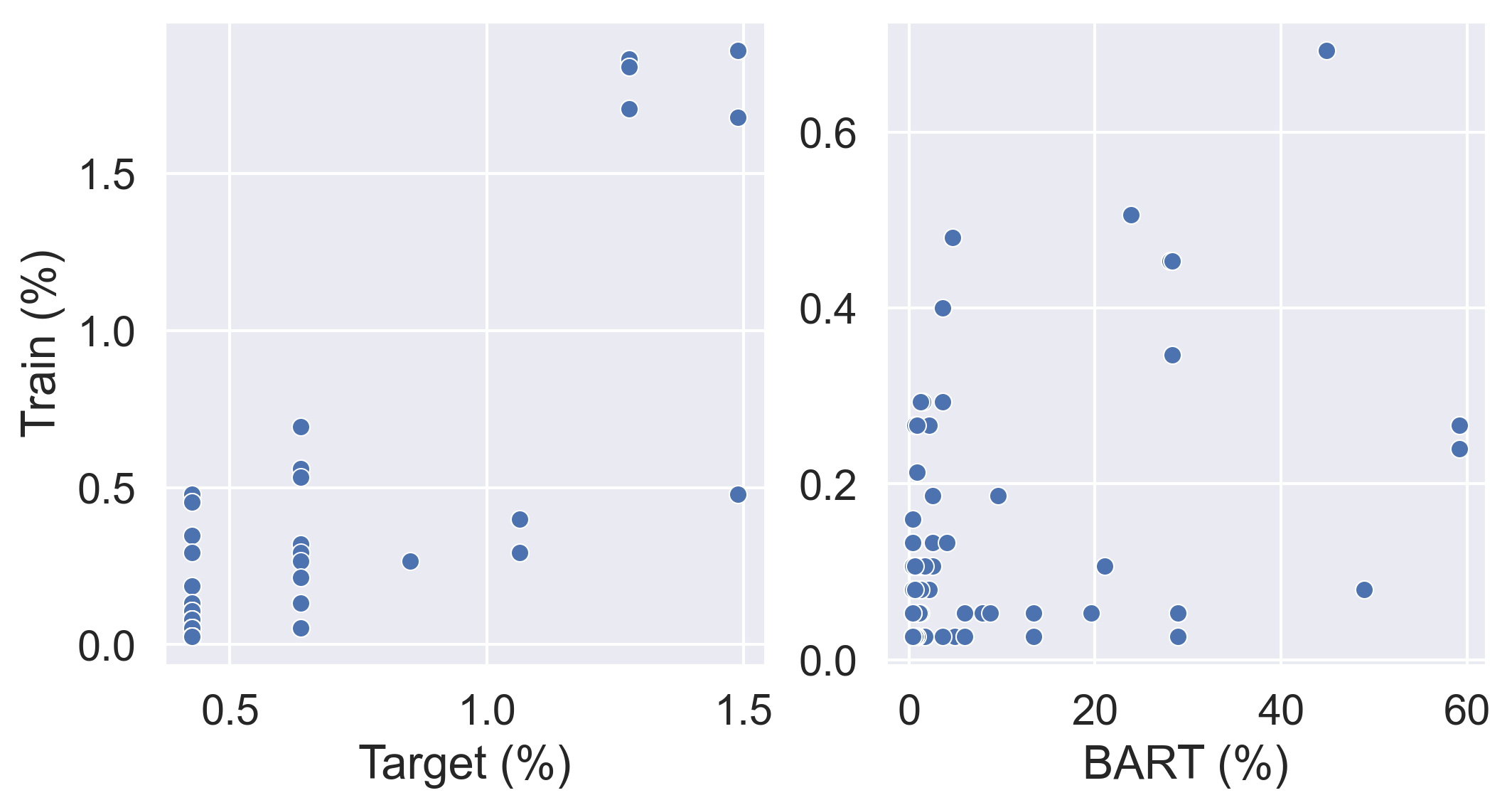}
\caption{Distribution of 7-grams in \textit{Train} vs.~\textit{Target} set summaries (left) and vs.~\textit{BART} summaries (right).}
\label{fig:train_vs_test}
\end{figure}

To determine the origin of these long $n$-grams, we calculate their overlap with summaries in the \emph{Train} set and their corresponding input documents. While overlap with inputs is nearly zero,  
up to 90\% of long $n$-grams are also found in \emph{Train} set summaries (Figure \ref{fig:overlap_train} in App.~\ref{app:repetition_in_train}). Interestingly, models with global attention (LED or PRIMERA-based) seem to replicate more long sequences from the \emph{Train} set summaries than BART-based ones, while in the Pegasus-based system (SciSpace) a smaller amount of self-repetition can be explained by fine-tuning. Finally, we observe that though the distributions of self-repeating $n$-grams in the target summaries of the \emph{Test} set and \emph{Train} set are very similar (Figure~\ref{fig:train_vs_test}; left), in generated summaries the rate of self-repetition increases up to 500x compared to occurrence in the \emph{Train} set summaries (Figure \ref{fig:train_vs_test}; right).
Models amplify repeating patterns from the \emph{Train} set to unnatural proportions!

\section{Automated evaluation metrics}
\label{sec:correlations}

We compute automated metrics for each generated summary and include instance-level scores in our dataset. We investigate how these metrics correlate with other metrics (\S\ref{sec:autoauto}) and with human evaluation facets (\S\ref{sec:autohuman}).

\paragraph{Metrics from the MSLR leaderboard:}
\begin{itemize}[itemsep=0pt, topsep=2pt, leftmargin=0pt]
    \item[] \textbf{ROUGE}: The leaderboard reported system-level ROUGE-1, ROUGE-2, and ROUGE-L F-scores \cite{lin-2004-rouge}. We report these same three metrics; in some plots, due to space constraints, we show the average of these three ROUGE metrics, which we call Avg-ROUGE-F.
    \item[] \textbf{BERTScore}: We compute and report BERTScore-F \cite{Zhang2020BERTScoreET} for each generated summary as computed using the RoBERTa-large model.
    \item[] \textbf{Delta-EI}: We compute Delta-EI as introduced by \citet{Wallace2020GeneratingN} and modified by \citet{DeYoung2021MS2MS} for the MSLR shared task. The metric computes the probability distributions of evidence direction for all intervention-outcome (I/O) pairs between inputs and the target and generated summaries. The final score is a sum over the Jensen-Shannon Divergence of probability distributions over all I/O pairs. 
    Lower values indicate higher similarity to the target summary. 
\end{itemize}

\paragraph{Other metrics we propose and examine:}
\begin{itemize}[itemsep=0pt, topsep=2pt, leftmargin=0pt]
    \item[] \textbf{NLI/STS/ClaimVer}: 
    These metrics leverage Sentence-BERT \cite{reimers-gurevych-2019-sentence} and are computed as the cosine similarity between the embedding of the target summary and the embedding of the generated summary when encoded with trained SBERT models. We use three pretrained variants of SBERT: RoBERTa fine-tuned on SNLI and MultiNLI (NLI); RoBERTa fine-tuned on SNLI, MultiNLI, and the STS Benchmark (STS); and PubMedBERT fine-tuned on MS-MARCO and the SciFact claim verification dataset (ClaimVer).
    \item[] \textbf{PIO-Overlap}: Following \citet{otmakhova-etal-2022-led}, we employ a strong PIO extractor (BioLinkBERT \citep{yasunaga-etal-2022-linkbert} trained on EBM-NLP \citep{nye-etal-2018-corpus}) to extract PIO spans. For each target-generated pair, we define PIO-Overlap as the intersection of the two extracted sets of PIO spans normalized by the number of PIO spans in the target summary. Spans are only considered to overlap if they have the same label and one span is a subspan of the other.
\end{itemize}

\subsection{Correlation between automated metrics}
\label{sec:autoauto}

We compute Pearson's correlation coefficients between pairs of metrics (Figure~\ref{fig:metric_correlations} in App.~\ref{app:corr_cochrane}). Most automated metrics are significantly correlated (p $<$ 0.01), except Delta-EI and PIO-Overlap. 
ROUGE and BERTScore show a strong positive correlation (r = 0.75), and NLI and STS have a strong positive correlation (r = 0.92), unsurprising since the underlying models are trained on similar data. Delta-EI presents as bimodal, with two peaks around 0 and 1. Distributions of instance-level automated metrics per system are shown in App.~\ref{app:autometrics}. 

System ranks (\S\ref{sec:rankings}) produced by automated metrics are highly correlated except for PIO-Overlap, which is anti-correlated (Figure~\ref{fig:spearman_all}). Ordering systems based on these metrics generally result in the same or similar rankings ($\rho$ $\geq$ 0.77 for all pairs of metrics besides PIO-Overlap), e.g., rankings from ClaimVer, NLI, and STS are identical ($\rho$ = 1).

\subsection{Correlation between automated metrics and human judgements}
\label{sec:autohuman}

We investigate the relationship between automated metrics and human facet-based annotations. For this analysis, we normalize human facets to 4 agreement scores: Fluency, PIO, Direction, and Strength, each in the range [0, 1] (details in App.~\ref{app:normalization}).

\begin{table}[t!]
    \footnotesize
    \centering
    \begin{tabular}{lcccc}
        \toprule
        Metric & Flu. & PIO & Dir. & Str. \\
        \midrule
        ROUGE & -0.014 & -0.010 & 0.007 & -0.035 \\  
        BERTScore & -0.000 & 0.022 & 0.036 & -0.033 \\
        Delta-EI & 0.066 & -0.080 & -0.060 & -0.054 \\
        ClaimVer & -0.051 & 0.142** & -0.017 & -0.093* \\  
        NLI & -0.026 & 0.053 & -0.011 & -0.063 \\
        STS & -0.042 & 0.066 & 0.001 & -0.056 \\
        PIO-Overlap & 0.043 & 0.358** & 0.033 & 0.050 \\
        \bottomrule
    \end{tabular}
\caption{Correlation coefficients between automated metrics and human evaluation facets. There is weak to no correlation between metrics and human-assessed facets (aside from between PIO-overlap and PIO). Statistical significance at $\alpha$ = 0.05 is marked with *, 0.01 with **, though these thresholds for significance do not account for multiple hypothesis testing.}
\label{tab:autohuman}
\end{table}

Correlation coefficients between automated metrics and these four agreement scores are given in Table~\ref{tab:autohuman}; PIO correlations are plotted in Figure~\ref{fig:ei_relation} in App~\ref{app:corr_cochrane}. In general, there is weak to no correlation between metrics and human-assessed Fluency, PIO, Direction, and Strength, suggesting that automated metrics may not be adequately capturing aspects of summaries that humans determine to be important. The exception is PIO-Overlap, which has a statistically significant correlation with human-assessed PIO agreement, and presents as a promising future metric for the MSLR task; ClaimVer is also weakly correlated with PIO agreement.

Disappointingly, Delta-EI does not correlate with human-assessed Direction agreement. 
We investigate this further by computing empirical cumulative distribution functions (ECDFs) for each of the metrics w.r.t.~Direction agreement (App.~\ref{app:corr_cochrane}). Delta-EI exhibits a small but desirable difference between instances where Direction agrees and instances where Direction disagrees (Agrees is more likely to have lower Delta-EI scores than Disagrees). 
In sum, Delta-EI shows some promise in detecting differences in Direction agreement, though further refinement of the metric is needed.

\section{Comparing system rankings}
\label{sec:rankings}

Evaluation metrics for summarization can be used in two settings, to judge performance at the \emph{instance} level (comparing individual summaries) or at the \emph{system} level (comparing model performance over many instances). Here, we compare system-level rankings produced by automated metrics, human facet evaluation, and pairwise preference annotations to determine whether automated metrics effectively rank systems as humans would. 

System rankings are computed by averaging the instance-level metric values or scores across all review instances for each system, and ranking from best to worst average score (direction depends on metric; higher is better for all scores except Delta-EI). We only average metrics over the subset of reviews for which we have human annotations. This ensures a fair comparison in the circumstance where we have selected an annotation sample that a system performs particularly well or poorly on. By doing this, the system rankings we present here are different than those computed using the same metrics from the MSLR leaderboards. We do not intend our computed rankings to be interpreted as the true system ranking; our analysis focuses on whether automated metrics and human evaluation are able to produce \emph{similar} rankings of systems. Table~\ref{tab:ranks} shows rankings as assessed by all automated metrics and human scores; Figure~\ref{fig:spearman_all} shows Spearman correlation coefficients. 

\begin{figure}[t!]
\includegraphics[width=\columnwidth]{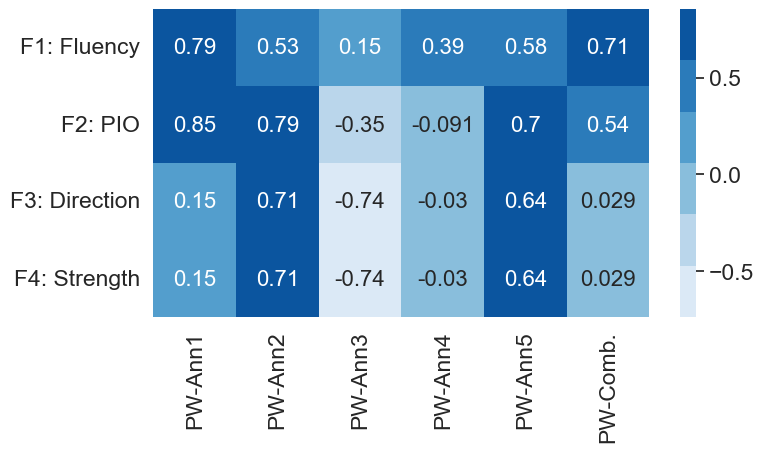}
\caption{Spearman rank correlations between system ranks for each pairwise annotator and ranks derived from facet-based annotation. Annotators weigh quality facets differently when performing pairwise judgments.}
\label{fig:pw_facets}
\end{figure}

\paragraph{Rankings by automated metrics are not correlated with rankings by human evaluation} In general, system rankings from commonly reported automated metrics are not correlated or anti-correlated (lighter blue) with system rankings produced by human judgments.
System rankings from automated metrics are highly correlated among themselves ($\rho$ close to 1), aside from PIO-Overlap. PIO-Overlap rankings are strongly correlated with rankings from human PIO agreement. PIO-Overlap and Delta-EI ranks also correlate with the combined pairwise rankings, again suggesting that these two metrics may be the most promising for capturing human notions of summary quality. 

\paragraph{Pairwise assessments do not weigh facets equally }
Pairwise-combined rankings are correlated with facet-based rankings for Fluency and PIO, but not Direction and Strength of claim. This may indicate that Fluency and PIO are more detectable problems, or that issues in Fluency and PIO are more prevalent in our data. The rank correlations also show that Direction and Strength are highly correlated and may capture similar aspects of system-level summary quality, making the case for dropping one of the two (likely Strength) in future annotations.

\paragraph{Pairwise preferences suggest that annotators weigh facets differently} In Figure~\ref{fig:pw_facets}, we show Spearman correlation coefficients of facet-based rankings against the rankings of five pairwise annotators and the combined pairwise ranking. These coefficients suggest that annotators weigh facets differently when comparing system output. Annotator 1 ranks similarly to Fluency and PIO facets, Annotators 2 and 5 rank similarly to PIO and Direction facets, while Annotators 3 and 4's rankings are uncorrelated with most facets. 

\section{Related work}
\label{sec:relatedwork}

Beyond ROUGE \cite{lin-2004-rouge} and BERTScore \cite{Zhang2020BERTScoreET}, an extensive list of $n$-gram \cite{papineni-etal-2002-bleu, banerjee-lavie-2005-meteor} and model-based \cite{zhao-etal-2019-moverscore, gao-etal-2020-supert, martins-etal-2020-sparse, sellam-etal-2020-bleurt, Yuan2021BARTScoreEG} summarization evaluation metrics have been proposed in the literature.
In particular, model-based approaches that use question generation and question answering \cite{wang-etal-2020-asking, durmus-etal-2020-feqa, deutsch-etal-2021-towards} or NLI-based models \cite{kryscinski-etal-2020-evaluating} have been proposed to assess summary factual consistency. \citet{fabbri-etal-2021-summeval} and \citet{deutsch-etal-2022-examining} provide more thorough evaluations of many of these metrics on select summarization tasks. 
We perform evaluations using metrics previously reported on the MSLR task, and leave a systematic evaluation of metrics on this task and others to future work.

In \citet{zhang-etal-2020-optimizing}, the authors performed fact verification on generated radiology reports using an information extraction module, by aligning the extracted entities with entities found in the reference summary. Our PIO-Overlap metric similarly uses a PIO entity extraction module to assess concept overlap between generated and reference summaries. 
\citet{falke-etal-2019-ranking} proposed to use NLI models to rank summaries by average entailment score per sentence against the input documents;
this shares similarities with the Delta-EI score we evaluated, which attempts to quantify agreement relative to the reference summary with respect to the direction of evidence reported.

\citet{deutsch-etal-2022-examining} investigated system-level rankings produced by automated metrics and human evaluation and found minimal correlation between them, 
a finding corroborated by our work. \citet{Liu2022RevisitingTG} introduced the robust summarization evaluation (RoSE) benchmark, containing human judgments for system outputs on the CNN/DM, XSum, and SamSum datasets. We extend such work into a novel domain (medical MDS for literature review) and demonstrate differences in automated metric performance and human evaluation in our domain and task. For example, though ROUGE correlates with human preferences in single-document (CNN/DM) and multi-document (MultiNews) news summarization, we find that it is poorly correlated with human judgments and preferences in the MSLR task. 

Recent developments in large language modeling have also shifted the goalposts for evaluation. \citet{Goyal2022NewsSA} found that although humans overwhelmingly prefer zero-shot GPT-3 summaries for news summarization, automated metrics were unable to capture this preference; they introduced a benchmark of human judgments and rationales comparing system outputs on the single-document news summarization task. More recently, \citet{Shaib2023SummarizingSA} demonstrated that GPT-3 can be adapted for the MSLR task, and though the model outputs are generally found by human annotators to be faithful to the inputs, in the MDS setting the evidence direction often disagrees with the reference. Detecting these disagreements and developing automated metrics that can capture such disagreements are valuable pursuits and one of the motivations for our work. Further investigation into whether automated metrics developed using limited human evaluation benchmarks such as the dataset we introduce here will be a goal for future work.

\section{Discussion}
\label{sec:discussion}

MDS for literature review may involve notions of summary quality not readily captured by standard summarization evaluation metrics. For example, our lexical analysis of generated summaries reveals a concerning level of self-repetition behavior, which is not penalized by standard metrics. Through two independent human evaluations (facet-based and pairwise preferences), we also show that automated metrics such as ROUGE and BERTScore are poorly correlated or even anti-correlated with human-assessed quality. 
This is not to say that these metrics do not provide any utility. Rather, further work is needed to understand what aspects of summary quality these metrics capture, and how to use them in combination with other metrics, novel metrics yet unintroduced, as well as human evaluation to better assess progress. We note that ours is not a systematic analysis of all automated summarization evaluation metrics, but is a focused study on evaluation metrics reported for the MSLR shared task and which we introduce under the hypothesis that they may be useful for capturing some quality facets associated with this task. For those interested in the former, please refer to studies such as \citet{fabbri-etal-2021-summeval} or \citet{deutsch-etal-2022-examining}.

A positive finding from our work is the promise of the PIO-Overlap and Delta-EI metrics. 
Delta-EI shows some potential to capture evidence directional agreement between summaries, though the metric as 
currently implemented is noisy and  
does not cleanly separate summaries that agree and disagree on direction. PIO-Overlap, a metric we introduce, correlates with human-assessed PIO agreement, 
suggesting that it could be a performant, scalable alternative to human evaluation of this quality facet. 
Still, more work is needed to probe how variants of these metrics could be adapted to evaluate performance on MSLR and other MDS tasks.

Finally, we note that human evaluation is difficult because people value different qualities in summaries. 
The rank-based analysis we perform does not account for interactions between related quality facets and is unable to elicit relationships between overall quality and individual quality facets. The majority of pairwise preference annotations in our dataset also include short free text justifications for preference decisions, which could be used to further study this problem. Other promising directions for future work involve studying how to optimally elicit human preferences, such as how to sample instances for labeling to maximize our confidence in the resulting system-level rankings.

\section{Conclusions}
\label{sec:conclusion}

There have been major recent advances in the generative capabilities of large language models. Models like ChatGPT,\footnote{\href{https://openai.com/blog/chatgpt}{https://openai.com/blog/chatgpt}} GPT-3 \cite{brown2020language}, and PubmedGPT\footnote{\href{https://hai.stanford.edu/news/stanford-crfm-introduces-pubmedgpt-27b}{https://hai.stanford.edu/news/stanford-crfm-introduces-pubmedgpt-27b}} demonstrate aptitude on many tasks but have also been shown to confidently produce factually incorrect outputs in specialized and technical domains.\footnote{Stack Overflow \href{https://meta.stackoverflow.com/questions/421831/temporary-policy-chatgpt-is-banned}{banned} ChatGPT responses due to the high rate of inaccurate and misleading information.}
Medicine is a specialized domain where incorrect information in generated outputs is difficult to identify and has the potential to do harm. There is therefore a pressing need for the community to develop better methods to assess the quality and suitability of generated medical texts. Our investigation confirms that there is significant room for improvement on medical MDS evaluation. We hope that the resources and findings we contribute in this work can assist the community towards this goal.

\section*{Limitations}
\label{sec:limitations}

Though we include 6 systems in our annotation which reflect the current state-of-the-art, all of the models are Transformer-based and fine-tuned on just the Cochrane dataset, which may limit the diversity of our generated summaries. Additionally, none of the systems are generating summaries that approach the accuracy of human-written summaries. As a consequence, though the summaries in our dataset span the spectrum of quality, they may have less coverage on the higher end of quality (summaries approaching the accuracy and utility of human-written review summaries).

Our analysis of evaluation metrics also assumes the existence of reference summaries. In many real-world summarization scenarios, reference summaries do not exist, and reference-free evaluation metrics are needed for assessment. We refer the reader to related work in reference-free summarization evaluation \cite{vasilyev-etal-2020-fill, gao-etal-2020-supert, luo-etal-2022-prefscore}, which have been found in some settings by \citet{fabbri-etal-2021-summeval} to exhibit even lower correlation with human notions of summary quality; the performance of these metrics on MSLR evaluation is unknown and is left to future work.

Our notions of summary quality also do not necessarily correspond to clinical utility. As with anything in the medical setting, it is of utmost importance to verify correctness and the quality of evidence before using any generated text to make or guide clinical decisions. 

\section*{Ethical Considerations}

As with other applications of NLP in the medical domain, results of MSLR systems must be verified by domain experts before they should be considered for use in clinical guidance. We do not intend the system outputs included in our dataset and analysis to be used for such end applications, as this would be clearly premature given the low quality of generated summaries and our lack of ability to assess the prevalence of factuality errors in these summary texts. Nonetheless, we believe that medical MDS holds eventual promise, and it is of vital importance that we study its challenges and how to measure and detect quality issues in generated text. 

\section*{Acknowledgements}

This research was partially supported by National Science Foundation (NSF) grant RI-2211954, and by the National Institutes of Health (NIH) under the National Library of Medicine (NLM) grant 2R01LM012086. YO and THT are supported by the Australian Government through the Australian Research Council Training Centre in Cognitive Computing for Medical Technologies (project number ICI70200030).

\bibliography{anthology,lucy}
\bibliographystyle{acl_natbib}

\appendix

\section{Facet-based Annotation}
\label{app:facet}

The questions and answer options shown to annotators for facet annotation are shown in Table~\ref{tab:app_facet}. If merging all Yes and Partial Yes classes, agreement proportion between annotators increases for Fluency (0.87 $\rightarrow$ 0.97), Population (0.56 $\rightarrow$ 0.64), Intervention (0.77 $\rightarrow$ 0.90), and Outcome agreement (0.36 $\rightarrow$ 0.44).

\begin{table*}[tbhp!]
    \footnotesize
    \centering
    \begin{tabular}{M{50mm}M{95mm}}
        \toprule
        Question & Answer options \\
        \midrule
        1. Is the generated summary fluent? & 
            \begin{itemize}[noitemsep, topsep=0pt, leftmargin=0pt]
                \item[] 2: Yes--there are no errors that impact comprehension of the summary
                \item[] 1: Somewhat, there are some minor grammatical or lexical errors, but I can mostly understand
                \item[] 0: No, there are major grammatical or lexical errors that impact comprehension 
            \end{itemize} \\
        \midrule
        2. Is the *population* discussed in the generated summary the same as the population discussed in the target summary? & 
            \begin{itemize}[noitemsep, topsep=0pt, leftmargin=0pt]
                \item[] 2: Yes
                \item[] 1: Partially
                \item[] 0: No
                \item[] N/A: No population in generated summary
                \item[] Other: Comment
            \end{itemize} \\
        \midrule
        3. Is the *intervention* discussed in the generated summary the same as the intervention discussed in the target summary? & 
            \begin{itemize}[noitemsep, topsep=0pt, leftmargin=0pt]
                \item[] 2: Yes
                \item[] 1: Partially
                \item[] 0: No
                \item[] N/A: No intervention in generated summary
                \item[] Other: Comment
            \end{itemize} \\
        \midrule
        4. Is the *outcome* discussed in the generated summary the same as the outcome discussed in the target summary? & 
            \begin{itemize}[noitemsep, topsep=0pt, leftmargin=0pt]
                \item[] 2: Yes
                \item[] 1: Partially
                \item[] 0: No
                \item[] N/A: No outcome in generated summary
                \item[] Other: Comment
            \end{itemize} \\
        \midrule
        5. What is the effect direction in the *target* summary for the main intervention and outcome considered? & 
            \begin{itemize}[noitemsep, topsep=0pt, leftmargin=0pt]
                \item[] (+1): Positive effect
                \item[] 0: No effect
                \item[] (-1): Negative effect
                \item[] N/A: no effect direction is specified in the target summary
                \item[] Other: Comment
            \end{itemize} \\
        \midrule
        6. What is the effect direction in the *generated* summary for the main intervention and outcome considered? & 
            \begin{itemize}[noitemsep, topsep=0pt, leftmargin=0pt]
                \item[] (+1): Positive effect
                \item[] 0: No effect
                \item[] (-1): Negative effect
                \item[] N/A: no effect direction is specified in the generated summary
                \item[] Other: Comment
            \end{itemize} \\
        \midrule
        7. What is the strength of the claim made in the *target* summary? & 
            \begin{itemize}[noitemsep, topsep=0pt, leftmargin=0pt]
                \item[] 3: Strong claim
                \item[] 2: Moderate claim
                \item[] 1: Weak claim
                \item[] 0: Not enough evidence (there is insufficient evidence to draw a conclusion)
                \item[] N/A: No claim (there is no claim in the summary)
                \item[] Other: Comment
            \end{itemize} \\
        \midrule
        8. What is the strength of the claim made in the *generated* summary? & 
            \begin{itemize}[noitemsep, topsep=0pt, leftmargin=0pt]
                \item[] 3: Strong claim
                \item[] 2: Moderate claim
                \item[] 1: Weak claim
                \item[] 0: Not enough evidence (there is insufficient evidence to draw a conclusion)
                \item[] N/A: No claim (there is no claim in the summary)
                \item[] Other: Comment
            \end{itemize} \\
        \bottomrule
    \end{tabular}
\caption{Questions and answer options used during facet annotation.}
\label{tab:app_facet}
\end{table*}

\section{Self-repetition rates in generated summaries}
\label{app:self_repetition}

Most of the long \textit{n}-grams repeating across documents contain meaningful statements regarding the direction or strength of effect findings rather than purely stylistic patterns, which means that the systems are prone to introducing factuality mistakes by replicating common statements. In Table~\ref{tab:selfrep_examples} we show the examples of the most repetitive 8-grams for the 6 models, together with the percentage of generated summaries they occur in.

\begin{figure}[t!]
\includegraphics[width=\columnwidth]{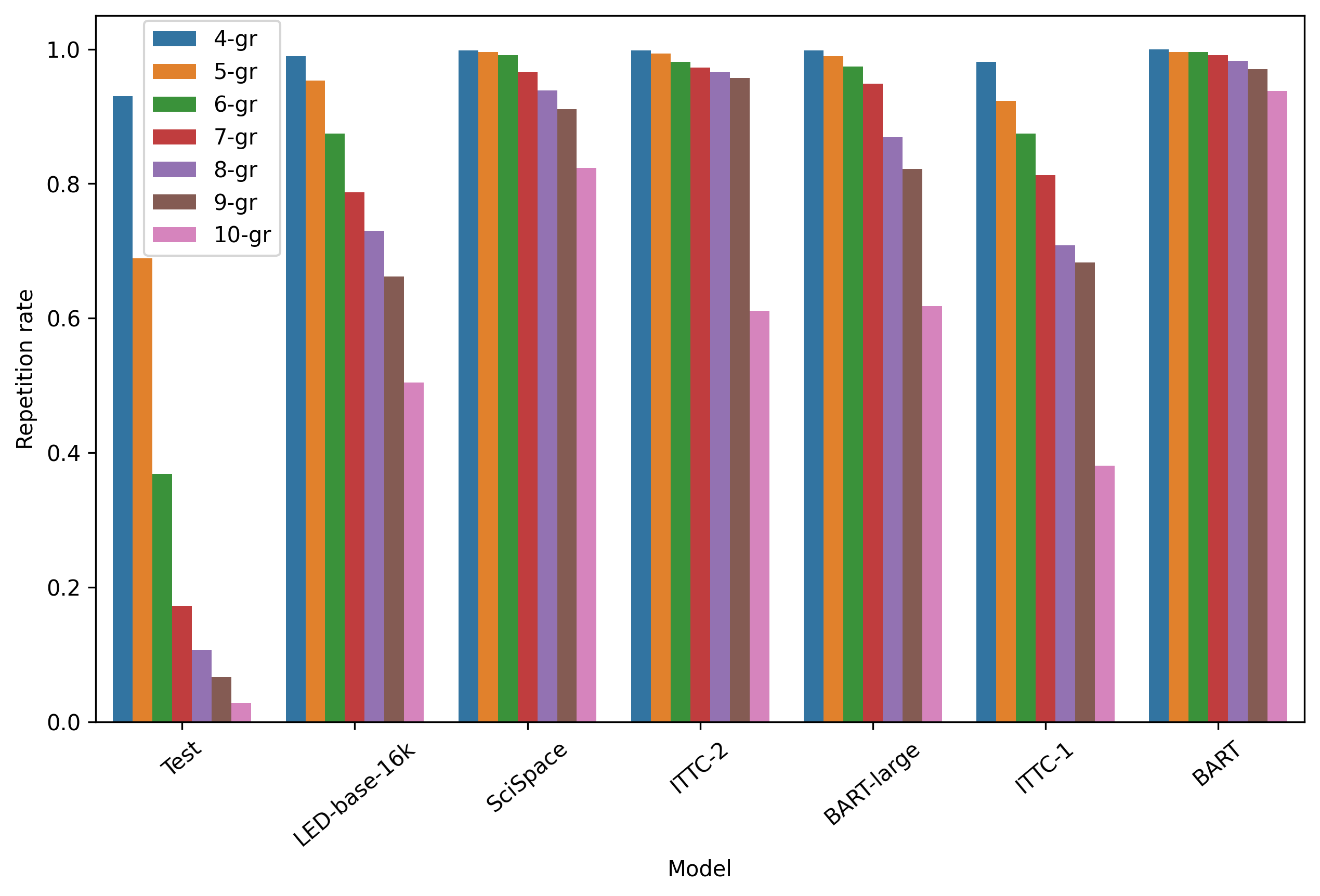}
\caption{Rate of self-repetition for models generations and the human written summaries (\textit{Test})}
\label{fig:cochraneselfrep}
\end{figure}

We also show that the self-repetition rate for $n$-grams with n > 4 have very dissimilar trends for generated summaries in comparison to human-written summaries (Figure \ref{fig:cochraneselfrep}) 
The amount of 5-grams and higher self-repetition also differs between models .

\begin{table*}[tbh!]
\footnotesize
\centering
\begin{tabular}{lcc}
\toprule
    Model & Most frequent 8-gram &  Self-repetition rate (\%) \\
    \midrule

    Targets  & \textit{the conclusions of the review once assessed .}    & 1.5 \\
    LED-base-16k      & \textit{there is insufficient evidence to support or refute} &   9.4                                      \\

    ITTC-1 & \textit{there is insufficient evidence to support the use} & 18.7 \\

    BART-large & \textit{there is insufficient evidence to support the use}  & 22.8 \\

    SciSpace       &  \textit{there is insufficient evidence to support the use} & 55.5  \\

    BART (baseline) & \textit{there is insufficient evidence from randomised controlled trials}  & 59.1 \\
    
    ITTC-2 &  \textit{there is insufficient evidence to support the use} & 65.1\\
\bottomrule
\end{tabular}

\caption{Examples of 8-grams which are most frequently repeated across generated summaries, together with their self-repetition rate.}
\label{tab:selfrep_examples}

\end{table*}

\begin{figure}[t!]
\includegraphics[width=\columnwidth]{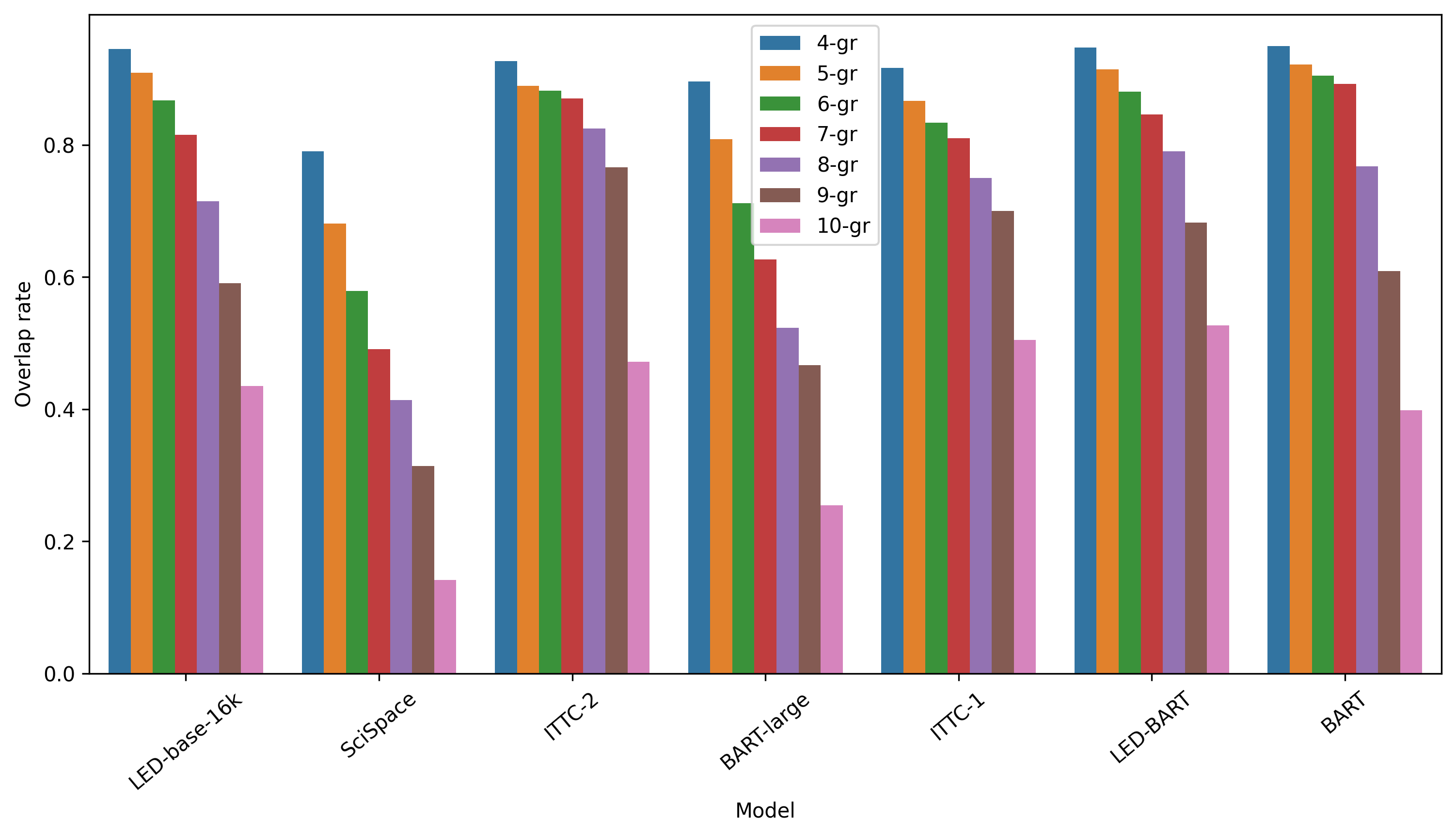}
\caption{Percentage of self-repeating $n$-grams in generated summaries which also occur in the target summaries of the Train set.}
\label{fig:overlap_train}
\end{figure}

\begin{figure*}[t!]
\includegraphics[width=\textwidth]{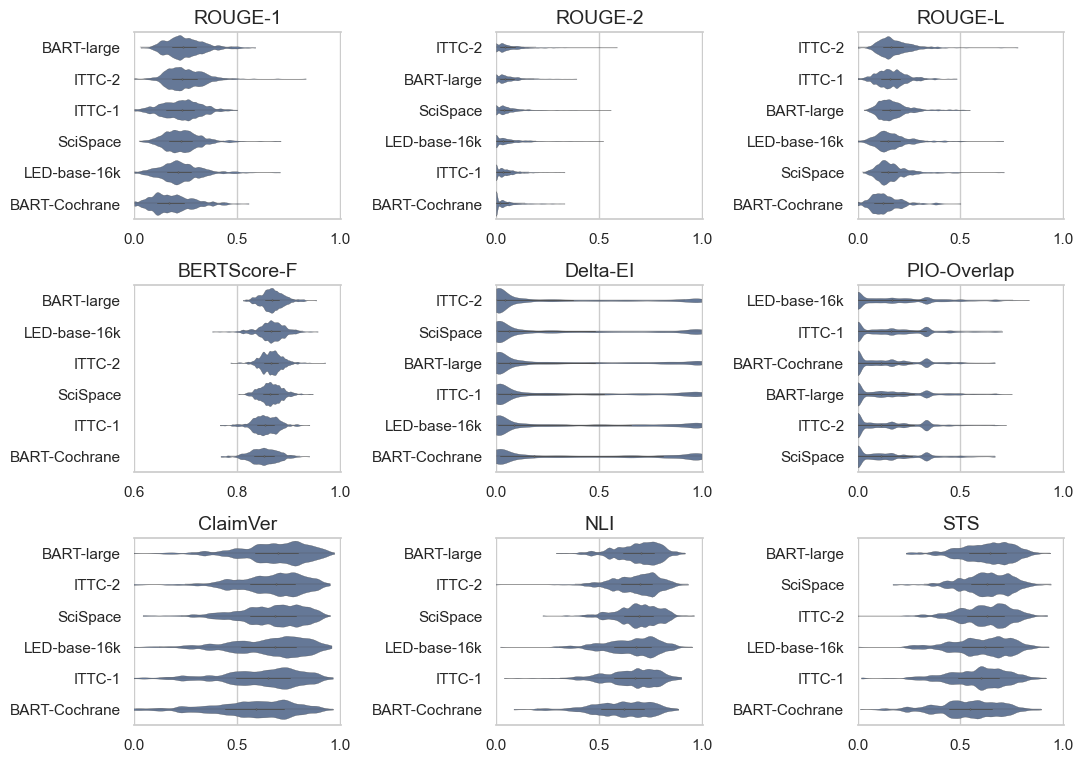}
\caption{Distribution of instance-level automated metric values by system (n=470 for each system). Each subplot is sorted on system from best to worst by median score. The median score ranking is typically not identical to the ranking of the corresponding metric from the MSLR leaderboard, which are computed based on micro-averaged metric values.}
\label{fig:metric_distro}
\end{figure*}

\section{Copying self-repeating $n$-grams from training set}
\label{app:repetition_in_train}

In Figure~\ref{fig:overlap_train}, we show the percentages of self-repeating $n$-grams from generated summaries which can also be found in the target summaries in the Train set.

\section{Automated metric distributions per system}
\label{app:autometrics}

Distributions of automated metrics for all instances per system are shown in Figure~\ref{fig:metric_distro}.

\section{Correlations between metrics in the Cochrane dataset}
\label{app:corr_cochrane}

\begin{figure*}[t!]
\centering
\includegraphics[width=0.75\textwidth]{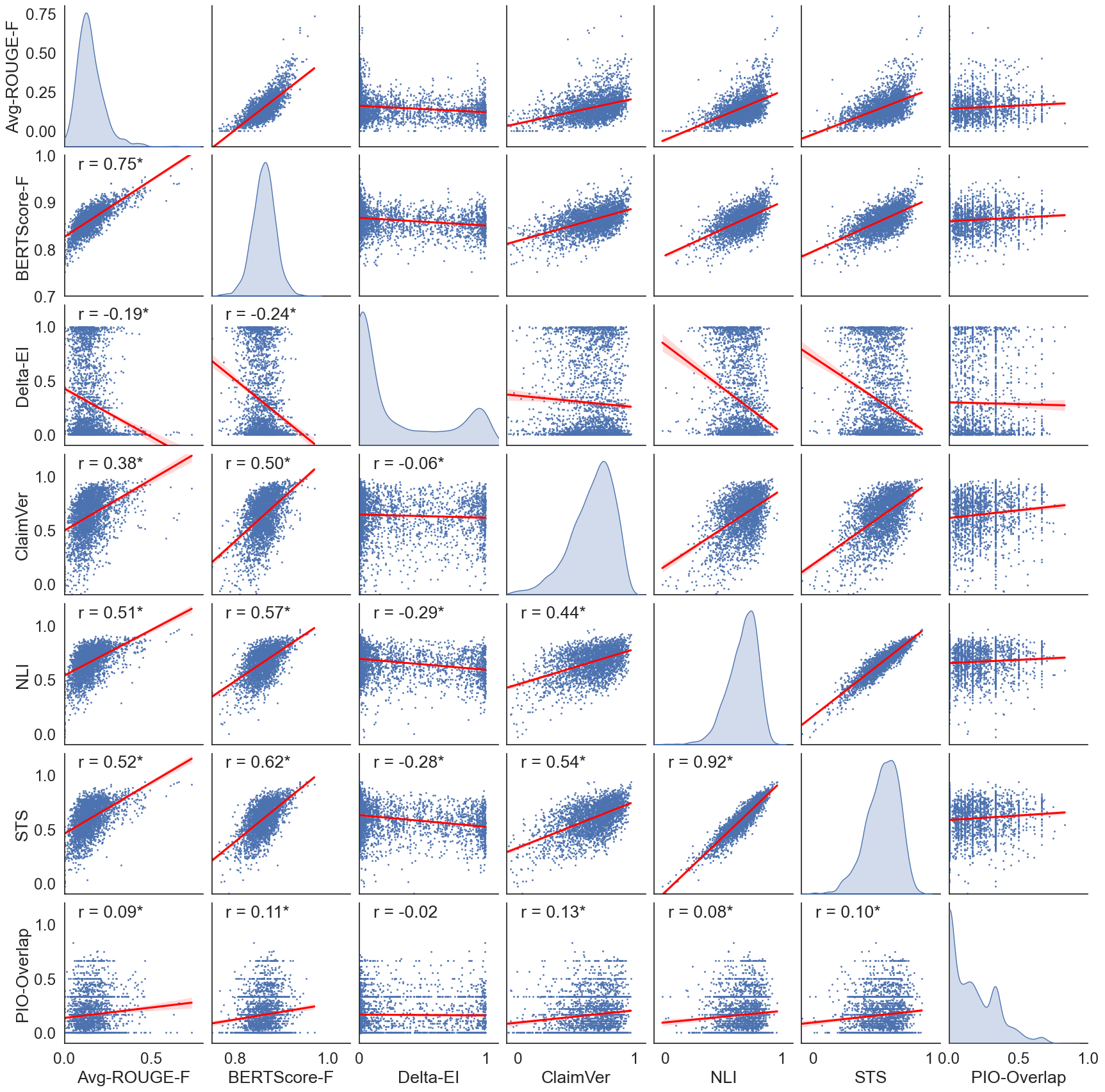}
\caption{Correlations between automated metrics in the Cochrane dataset. Pearson's correlation coefficients (r) are shown, along with an * if p < 0.01.}
\label{fig:metric_correlations}
\end{figure*}

We present correlations between all automated metrics along with correlation coefficients (Figure~\ref{fig:metric_correlations}). ROUGE and BERTScore are strongly correlated. NLI and STS are strongly correlated. Delta-EI has a bimodal distribution. PIO-Overlap is uncorrelated with other metrics. 

\begin{figure*}[t!]
\includegraphics[width=\textwidth]{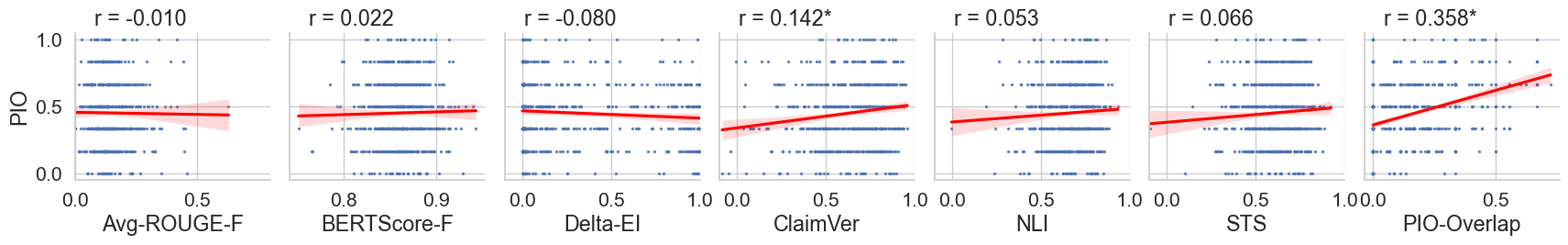}
\caption{Correlations between automated metrics and the normalized PIO facet score.}
\label{fig:pio_corr}
\end{figure*}

Correlations between automated metrics and the normalized PIO facet score are shown in Figure~\ref{fig:pio_corr}. In general, automated metrics are poor predictors of PIO agreement, except PIO-Overlap, which is positively correlated with PIO agreement (p < 0.05). This confirms that model extraction and alignment of PIO spans is a promising direction for assessing PIO agreement. ClaimVer also shows a weak but statistically significant correlation with PIO agreement. The ClaimVer metric is computed based on embedding similarity between two texts using a model trained on the SciFact scientific claim verification dataset \cite{wadden-etal-2020-fact}; the SciFact task measures whether evidence entails or refutes a scientific claim, which is somewhat analogous to our evaluation task for medical multi-document summarization.

\begin{figure*}[t!]
\includegraphics[width=\textwidth]{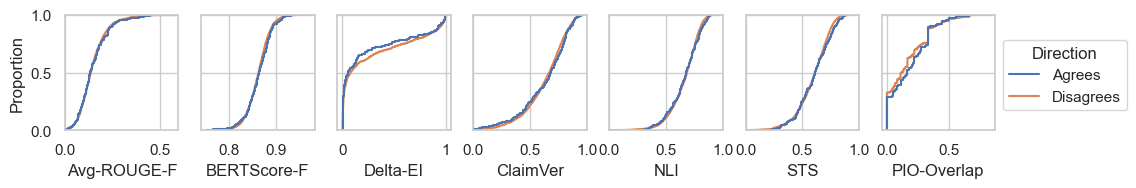}
\caption{Empirical cumulative distribution function (ECDF) of each of the automated metrics and their values for summaries where humans assessed the evidence direction to Agree versus those assessed to Disagree.}
\label{fig:ei_relation}
\end{figure*}

We also assess whether metrics can distinguish between summaries where the Direction agrees with the target and summaries where the Direction disagrees. We present the empirical cumulative distribution functions (ECDF) for each automated metric, showing the separation of metrics between when Direction agrees and disagrees (Figure~\ref{fig:ei_relation}. The Delta-EI metric is somewhat sensitive to human-assessed directional agreement (a higher proportion of generated summaries where the Direction agrees with the target have lower Delta-EI scores), though we note that the difference is small. PIO-Overlap also shows some separation between the two Direction classes (a higher proportion of disagrees have lower PIO-Overlap score than agrees), though again the difference is subtle.

\section{Normalizing human facet scores}
\label{app:normalization}

Responses to the Fluency question result in a 3-class ordinal variable that we map to the range [0, 1], where 0.0 is disfluent, 0.5 is somewhat fluent, and 1.0 is fluent. PIO aggregates agreement over Population, Intervention, and Outcome, where each of P, I, and O are 3-class ordinal variables that we map to the range [0, 1] as we do Fluency; we average the three facets to get PIO agreement. For evidence direction, though each of the two annotated questions has 4 answers (positive, no effect, negative, or no direction given), we elect to define Direction as a binary class. We normalize Direction to 1 if the target direction and generated direction agree and 0 if they disagree. For Strength, each of the two annotated questions has 4 answers (strong, moderate, weak, and not enough evidence). We take the difference between the answers for the target and generated summaries and normalize to the range [0, 1] to yield our Strength agreement score.

\end{document}

%% file: commands.tex
\newcommand\lucy[1]{{\color{blue}\{\textit{#1}\}$_{lucy}$}}

\newcommand\mstoo{MS\^{}2\xspace}

\newcommand\todoit[1]{{\color{red}\{TODO: \textit{#1}\}}}
\newcommand\todo{{\color{red}{TODO}}\xspace}
\newcommand\todocite{{\color{red}{CITE}}\xspace}

\newcommand\semanticscholar{{\small{\texttt{anon\_corpus}}}\xspace}
\newcommand\allenai{{\small{\texttt{anonymized}}}\xspace}
\newcommand\githublink{\href{https://github.com/allenai/mslr-annotated-dataset}{https://github.com/ allenai/mslr-annotated-dataset}\xspace}

\newcolumntype{L}[1]{>{\raggedright\let\newline\\\arraybackslash\hspace{0pt}}p{#1}}
\newcolumntype{M}[1]{>{\raggedright\let\newline\\\arraybackslash\hspace{0pt}}m{#1}}

\newcommand{\rulesep}{\unskip\ \textcolor{gray}{\vrule}\ }

\definecolor{darkgreen}{rgb}{0.0, 0.4, 0.13}